\documentclass[letterpaper, 10 pt, conference]{ieeeconf}  

\IEEEoverridecommandlockouts                              

\usepackage{amsmath,amsfonts}
\usepackage{algorithmicx}
\usepackage{algorithm,algpseudocode}
\usepackage{array}
\usepackage[caption=false,font=normalsize,labelfont=sf,textfont=sf]{subfig}
\usepackage{textcomp}
\usepackage{stfloats}
\usepackage{url}
\usepackage{verbatim}
\usepackage{graphicx}
\usepackage[percent]{overpic}
\usepackage{cite}
\usepackage[table]{xcolor}
\usepackage{booktabs}

\graphicspath{ {Images/} } 

\newcommand{\real}{\mbox{\rm I$\!$R}}

\newcommand{\bld}[1]{\mbox{\boldmath $#1$}} 

\usepackage{cuted}

\usepackage{xcolor}

\title{\LARGE \bf
Full-Dynamics Real-Time Nonlinear Model Predictive Control of Heavy-Duty Hydraulic Manipulator for Trajectory Tracking Tasks
}

\author{Alvaro Paz, Mahdi Hejrati, Pauli Mustalahti, and Jouni Mattila
\thanks{This work was supported by the Business Finland partnership project ''Future all-electric rough terrain autonomous mobile manipulators (Grant \#2334/31/2022)''. Corresponding author: Pauli Mustalahti.}
\thanks{All authors are with the Faculty of Engineering and Natural Sciences, Tampere University, Tampere, Finland.
       {\tt\footnotesize alvaro.pazanaya@tuni.fi, mahdi.hejrati@tuni.fi, pauli.mustalahti@tuni.fi, jouni.mattila@tuni.fi}}
}

\begin{document}

\maketitle
\thispagestyle{empty}
\pagestyle{empty}

\begin{abstract}
Heavy-duty hydraulic manipulators (HHMs) operate under strict physical and safety-critical constraints due to their large size, high power, and complex nonlinear dynamics. Ensuring that both joint-level and end-effector trajectories remain compliant with actuator capabilities—such as force, velocity, and position limits—is essential for safe and reliable operation, yet remains largely underexplored in real-time control frameworks. This paper presents a nonlinear model predictive control (NMPC) framework designed to guarantee constraint satisfaction throughout the full nonlinear dynamics of HHMs, while running at a real-time control frequency of 1~kHz. The proposed method combines a multiple-shooting strategy with real-time sensor feedback, and is supported by a robust low-level controller based on virtual decomposition control (VDC) for precise joint tracking. Experimental validation on a full-scale hydraulic manipulator shows that the NMPC framework not only enforces actuator constraints at the joint level, but also ensures constraint-compliant motion in Cartesian space for the end-effector. These results demonstrate the method’s capability to deliver high-accuracy trajectory tracking while strictly respecting safety-critical limits, setting a new benchmark for real-time control in large-scale hydraulic systems.

\end{abstract}
\section{INTRODUCTION}
Heavy-duty hydraulic manipulators (HHMs) play a vital role in a range of demanding industries, including mining, agriculture, forestry, offshore operations, and field robotics. Their widespread adoption is largely attributed to their superior power-to-weight ratio, which enables them to handle significantly heavier payloads compared to their electric counterparts. This capability allows HHMs to perform tasks that far exceed human physical limits. Due to their large size and substantial weight, HHMs are typically operated in two principal modes: predefined motion paths~\cite{hejrati2025orchestrated} and teleoperation~\cite{hejrati2025robust}. In both scenarios, the effective execution of commands necessitates a control system that is not only robust but also capable of maintaining high performance under varying operational conditions. As a result, the development of advanced control algorithms for HHMs has become a critical focus, particularly those that ensure both precision and resilience in complex industrial environments~\cite{mattila2017survey}.

Despite the numerous advantages of HHMs, they present significant control challenges due to their structural complexities, nonlinear actuator dynamics governed by fluid power systems, unmodeled uncertainties, and susceptibility to external disturbances~\cite{mattila2017survey}. These challenges are further amplified in real-time applications, where both robustness and computational efficiency are critical requirements. Recent years have witnessed a growing interest in advanced control strategies for HHMs. In~\cite{ortiz2014increasing}, a PID control approach augmented with a high-gain observer was developed to control a forestry crane. Similarly,~\cite{feng2018robotic} proposed a PID-based control scheme for excavator automation, where the control gains were optimized using a genetic algorithm. Reinforcement learning (RL) methods have also emerged as promising tools; for example,~\cite{egli2022general} applied RL to enhance automation capabilities in hydraulic excavators, while~\cite{taheri2024end} employed RL for end-effector velocity control of a redundant hydraulic crane. Alternatively, virtual decomposition control (VDC)~\cite{zhu2010virtual} has demonstrated notable success in HHM control by achieving sub-centimeter precision and robust performance in the presence of real-world disturbances and modeling uncertainties \cite{hejrati2025orchestrated}. Moreover, VDC has been shown to maintain high computational efficiency, making it suitable for real-time deployment in industrial environments~\cite{koivumaki2019energy, hejrati2025orchestrated}.

Although the control of HHMs has recently gained attention, safety-aware control strategies remain significantly underexplored. Given the size and power of these machines, any violation of state or actuator constraints—such as velocity or force limits—can lead to severe system damage and safety hazards. One widely adopted strategy to enforce kinematic and dynamic constraints in electro-hydraulic or hydraulic systems involves designing constraint-satisfying control laws~\cite{helian2023constrained,xu2022eso,xu2022extended}. Another promising direction is the use of optimization-based control approaches, such as model predictive control (MPC)~\cite{garcia1989model}, which can explicitly account for both system dynamics and constraints. A major trade-off in MPC design lies between linear and nonlinear formulations. Linear MPC offers computational efficiency but often falls short in handling the complex nonlinear dynamics inherent in hydraulic systems. In contrast, nonlinear MPC (NMPC) provides greater fidelity and constraint satisfaction but typically incurs high computational costs, posing challenges for real-time deployment~\cite{diehl2009efficient}.

Several recent studies have attempted to bridge this gap. In~\cite{varga2019model}, a linearized model of a heavy-duty machine was used to design an MPC controller, with numerical evaluation. A data-driven NMPC was proposed in~\cite{ma2024data}, employing a machine learning-based prediction model and implemented on a 22-ton system with a 50~Hz sampling rate. Additionally,~\cite{kalmari2014nonlinear} developed a NMPC controller with sway compensation for a real hydraulic forestry crane, achieving an average control cycle time of 37.6~ms. These results demonstrate the potential of NMPC for HHMs in satisfying safety and performance requirements. However, due to the high-DoF nonlinear dynamics of HHMs, existing implementations either rely on simplified models or operate at relatively low frequencies. To the best of our knowledge, no prior work has successfully implemented NMPC on the complete nonlinear dynamics of HHMs at a control frequency of 1~kHz—despite such frequencies being demonstrated on smaller-scale robotic platforms~\cite{kleff2021high}. 

This research lays a critical foundation for the future integration of learning-based skill transfer methods in HHMs. The proposed NMPC framework is designed not only as a high-performance controller, but also as a safety-critical interface between high-level decision-making—driven by learning algorithms—and the low-level actuation handled by VDC. By ensuring that learned commands are translated into constraint-compliant, executable trajectories, our approach enables the safe deployment of intelligent behaviors in real-world and industrial environments. In doing so, it bridges the gap between data-driven autonomy and the strict physical limitations of large-scale hydraulic systems—paving the way for the next generation of intelligent, safe, and adaptive heavy machinery.




\subsection{Contributions}

The multiple-shooting algorithm introduced in~\cite{bock1984multiple} enables the development of NMPC frameworks that offer both robust and optimal control performance with real-time feasibility~\cite{diehl2006fast}. In this work, we consider the full analytic nonlinear dynamics of the hydraulic manipulator, capturing all relevant physical and actuator-level nonlinearities. We present a novel NMPC framework designed to accurately track reference trajectories while meeting the strict computational demands of real-time execution. The proposed method incorporates several key components to achieve these goals. First, we integrate a multiple-shooting strategy informed by high-frequency kinematic sensor measurements, which improves estimation accuracy. Second, a robust low-level controller based on VDC is employed to ensure precise tracking of joint positions and velocities, complementing the NMPC high-level strategy.

To address real-time constraints, the proposed control framework leverages a combination of techniques that collectively reduce computation time and ensure deterministic execution. These include: 
\begin{itemize}
    \item Warm-starting of the solver using previous solutions,
    \item High-frequency sensor sampling at 1~kHz,
    \item A bounded maximum number of iterations within the solver,
    \item Efficient sensor buffer management routines to prevent data saturation and reduce computational overhead.
\end{itemize}

These design choices ensure that the optimal control problem is solved within the sensor sampling interval, guaranteeing that control inputs are computed before the arrival of the next measurement. The resulting deterministic computation time is a key requirement for real-time control, and it is successfully achieved in our implementation.




\subsection{Paper Outline}

The remainder of this paper is organized as follows. Section~\ref{sec:nlp} presents the formulation of the nonlinear programming (NLP) problem and the definition of the cost function. Section~\ref{sec:nmpc} details the proposed nonlinear model predictive control (NMPC) algorithm. Experimental results and performance evaluations are provided in Section~\ref{sec:results}. Finally, Section~\ref{sec:conclusion} concludes the paper and outlines directions for future work.

\section{Nonlinear Programming Problem}
\label{sec:nlp}
A parallel-serial heavy duty manipulator is composed by serial and closed-kinematics chains modules \cite{alvaro2024analytical} and its dynamics can be expressed in generalized coordinates as
\begin{equation}
    \bld{f}\!_{x} \ = \ \bld{H}(\bld{\theta})\,\ddot{\!\bld{\theta}} + \bld{h}(\bld{\theta},\,\dot{\!\bld{\theta}})
    \label{eq:991}
\end{equation}
where $\bld{\theta}\in\real^n$ is the configuration vector and $n$ is the number of degrees of freedom. The vectors $\,\dot{\!\bld{\theta}},\,\ddot{\!\bld{\theta}}\in\real^n$ are the generalized joint velocity and acceleration, respectively; and $\bld{f}\!_{x}\in\real^n$ is the generalized joint forces/torques. The symmetric positive-definite matrix $\bld{H}(\cdot)\in\real^{n\times n}$ is the generalized inertia matrix and $\bld{h}(\cdot)\in\real^n$ is the vector that contains gravitational, centrifugal and nonlinear effects. Expression (\ref{eq:991}) represents the inverse dynamics problem which efficient solutions have been reported \cite{petrovic2022mathematical} but in order to implement a multiple shooting approach \cite{bock1984multiple,diehl2006fast} it is required to compute the forward dynamics solution, i.e.
\begin{equation}
    \,\ddot{\!\bld{\theta}} \ = \ \bld{H}(\bld{\theta})^{-1} \left( \bld{f}\!_{x} - \bld{h}(\bld{\theta},\,\dot{\!\bld{\theta}}) \right)
    \label{eq:992}
\end{equation}
which analytical solutions have been reported in \cite{alvaro2024analytical} by relying on geometric recursive algorithms based on screw theory. Then this solution is expressed by the analytic function
\begin{equation}
    \,\ddot{\!\bld{\theta}} \ = \ \bld{F\!D}\,(\bld{\theta},\,\dot{\!\bld{\theta}},\bld{f}\!_{x})
    \label{eq:993}
\end{equation}

Additionally, we define the robot's TCP pose and twist as $\bld{x}(t)$ and $\dot{\bld{x}}(t)$, respectively. Such terms can be computed through forward kinematics by knowing $\bld{\theta}$ and $\,\dot{\!\bld{\theta}}$.

Thus, let $\bld{x}^{r}(t)$ be a continuous-and-differentiable time $t$ reference trajectory on the cartesian working space of the robot and $\dot{\bld{x}}^{r}(t)$ its first time derivative. Thus, an infinite dimensional constrained optimal control problem (OCP) for tracking $\bld{x}^{r}(t)$ and $\dot{\bld{x}}^{r}(t)$ with minimum effort is defined as
\begin{eqnarray}
\hspace*{-0.65cm}\underset{\bld{\theta} (t), \ \,\dot{\!\bld{\theta}} (t)}{\operatorname{min}} & \!\!\! & \tfrac{1}{2} \int_{0}^{T} \bld{f}\!_{x}(t)^{\!\top} \bld{f}\!_{x}(t) \ dt
\label{eq:costf1} \\
\nonumber \\
\hspace*{-0.65cm} \mbox{subject to} & \!\!\! &
\left\{\begin{array}{lcl}
\bld{f}\!_{x} & = & \bld{H}(\bld{\theta})\,\ddot{\!\bld{\theta}} + \bld{h}(\bld{\theta},\,\dot{\!\bld{\theta}}) \vspace*{-0.0cm} \\
\bld{x}(t) & = & \bld{x}^r (t) \vspace*{-0.0cm} \\
\dot{\bld{x}}(t) & = & \dot{\bld{x}}^r (t) \vspace*{-0.0cm} \\
\bld{x}_{{\scriptsize \mbox{min}}} & \leq & \bld{x}(t) \ \ \leq \ \ \ \bld{x}_{{\scriptsize \mbox{max}}} \vspace*{-0.0cm} \\
\bld{\theta}_{{\scriptsize \mbox{min}}} & \leq & \ \bld{\theta}(t) \ \ \leq \ \ \ \bld{\theta}_{{\scriptsize \mbox{max}}} \vspace*{-0.0cm} \\
\dot{\bld{x}}_{{\scriptsize \mbox{min}}} & \leq & \dot{\bld{x}}(t) \ \ \leq \ \ \ \dot{\bld{x}}_{{\scriptsize \mbox{max}}} \vspace*{-0.0cm} \\
\dot{\!\bld{\theta}}_{{\scriptsize \mbox{min}}} & \leq & \ \dot{\!\bld{\theta}}(t) \ \ \leq \ \ \ \dot{\!\bld{\theta}}_{{\scriptsize \mbox{max}}} \vspace*{-0.0cm} \\
\bld{f}\!_{x\,{\scriptsize \mbox{min}}} & \leq & \bld{f}\!_{x}(t) \ \ \leq \ \ \ \bld{f}\!_{x\,{\scriptsize \mbox{max}}} \vspace*{-0.0cm} \\
\end{array} \right. \ ,
\label{eq:const1}
\end{eqnarray}
where the cost function minimizes the actuators' effort over the time $t\in[0,T]$. The constraints stand for the system dynamics (\ref{eq:991}) and the reference trajectory tracking. The robot's TCP motion is bounded by the minimum limits $\bld{x}_{{\scriptsize \mbox{min}}}$ and $\dot{\bld{x}}_{{\scriptsize \mbox{min}}}$ and maximum limits $\bld{x}_{{\scriptsize \mbox{max}}}$ and $\dot{\bld{x}}_{{\scriptsize \mbox{max}}}$. The joint limits are bounded by $\bld{\theta}_{{\scriptsize \mbox{min}}}$ and $\bld{\theta}_{{\scriptsize \mbox{max}}}$ in position and by $\dot{\!\bld{\theta}}_{{\scriptsize \mbox{min}}}$ and $\dot{\!\bld{\theta}}_{{\scriptsize \mbox{max}}}$ in velocity. Also, the actuators' effort is bounded by the minimum and maximum limits $\bld{f}\!_{x\,{\scriptsize \mbox{min}}}$ and $\bld{f}\!_{x\,{\scriptsize \mbox{max}}}$.

The process of converting an infinite-dimensional OCP into a finite-dimensional nonlinear programming problem (NLP) \cite{Bib:Betts} is known as transcription, shuch process starts by discretize the time as $t_k \, \forall \, k\in\{1 \!\cdots\! N\}$ where $N\in \mathbb{Z}_{>0}$ is the time horizon, then the time increment is ${\Delta_t} = t_{k+1}-t_k$. By defining the joint states and effort at instant time $k$ as $\bld{\theta}_{k}$, $\,\dot{\!\bld{\theta}}_{k}$ and $\bld{f}\!_{xk}$ we can evaluate (\ref{eq:991}) and retrieve the joint acceleration at time $k$ as $\,\ddot{\!\bld{\theta}}_{k} = \bld{F\!D}\,(\bld{\theta}_{k},\,\dot{\!\bld{\theta}}_{k},\bld{f}\!_{xk})$. Then the joint states at instant time $k\!+\!1$ can be approximated by
\begin{equation}
    [ \,\hat{\!\bld{\theta}}_{k\!+\!1}, \ \,\dot{\hat{\!\bld{\theta}}}_{k\!+\!1} ] \ = \ \text{integrator}( \bld{\theta}_{k}, \ \,\dot{\!\bld{\theta}}_{k}, \ \,\ddot{\!\bld{\theta}}_{k}, \ \Delta_t )
    \label{eq:int}
\end{equation}
where the $\text{integrator}( \cdot )$ function can perform numeric integration techniques, e.g. Euler step, Runge Kutta and trapezoidal. The states approximations at $k\!+\!1$ are denoted by $\,\hat{\!\bld{\theta}}_{k\!+\!1}$ and $\,\dot{\hat{\!\bld{\theta}}}_{k\!+\!1}$ for position and velocity, respectively.

Let us transcribe now the aforementioned OCP (\ref{eq:costf1}-\ref{eq:const1}) into a NLP that describes our NMPC, thus we roll out the discrete states and controls over the time horizon and define the following decision variable $\bld{z}\in\real^{3nN}$
\begin{equation}
    \bld{z} \ = \ [ \ \bld{f}\!_{x1}^{\top} \ \bld{\theta}_{1}^{\top} \ \,\dot{\!\bld{\theta}}_{1}^{\top} \ \cdots \ \bld{f}\!_{xk}^{\top} \ \bld{\theta}_{k}^{\top} \ \,\dot{\!\bld{\theta}}_{k}^{\top} \ \cdots \ \bld{\theta}_{N}^{\top} \ \,\dot{\!\bld{\theta}}_{N}^{\top} \ ]^{\top}
\end{equation}
then at time $t_k \, \forall \, k\in\{1 \!\cdots\! N\!-\!1\}$ we define a running cost
\begin{eqnarray}
    L_k ( \bld{\theta}_{k}, \,\dot{\!\bld{\theta}}_{k}, \bld{f}\!_{xk}, \bld{x}_{k}^{r}, \dot{\bld{x}}_{k}^{r} ) & \!\!\!=\!\!\! & \left\| \bld{x}_{k} \!-\! \bld{x}_{k}^{r} \right\|_{Q_x}^2 +\left\| \dot{\bld{x}}_{k} \!-\! \dot{\bld{x}}_{k}^{r} \right\|_{Q_{\dot{x}}}^2 \nonumber \\
    &&+ || \bld{f}\!_{xk} ||_R^2
\end{eqnarray}
where $\bld{x}_{k}^{r}$ is the reference trajectory at time $t_k$ and $\dot{\bld{x}}_{k}^{r}$ is its time derivative. The objectives apply the weighted quadratic norm $|| \cdot ||^2$ where $R\in\real^{n\times n}$ is the weighting matrix that minimize the input control and the matrices $Q_x , Q_{\dot{x}}\in\real^{n\times n}$ penalize the tracking deviation in position and velocity.

In a similar way, the final cost at time $t_N$ is defined by
\begin{equation}
    L_N ( \bld{\theta}_{\!N}, \,\dot{\!\bld{\theta}}_{\!N}, \bld{x}_{\!N}^{r}, \dot{\bld{x}}_{\!N}^{r} ) = \left\| \bld{x}_{\!N} \!-\! \bld{x}_{\!N}^{r} \right\|_{Q_x}^2 + \left\| \dot{\bld{x}}_{\!N} \!-\! \dot{\bld{x}}_{\!N}^{r} \right\|_{Q_{\dot{x}N}}^2
\end{equation}
where the weighting matrices $Q_{x\!N} , Q_{\dot{x}\!N}\in\real^{n\times n}$ penalize the same deviation at $t_N$ for position and velocity.

A state feedback controller can thus be defined by the following optimization problem considering the current ($t_0$) state variable measurements as $\bld{\theta}_{{\scriptsize \mbox{msr}}}$ and $\,\dot{\!\bld{\theta}}_{{\scriptsize \mbox{msr}}}$
\begin{equation}
	\hspace*{-0.17cm}\underset{\bld{z}}{\text{arg min}}  \ \ \ J \ = \ \tfrac{1}{2} \sum_{k=1}^{N-1} \hspace{-0.05cm} L_k ( \cdot ) + L_{\!N} ( \cdot )
	\label{eq:costf2}
\end{equation}
\hspace{0.0cm}subject to
\begin{subequations}
\begin{eqnarray}
\bld{\theta}_{0} & = & \bld{\theta}_{{\scriptsize \mbox{msr}}} \ \qquad \text{and} \ \qquad \,\dot{\!\bld{\theta}}_{0} \ = \ \,\dot{\!\bld{\theta}}_{{\scriptsize \mbox{msr}}} \label{eq:res_a} \\
\bld{\theta}_{k\!+\!1} & = & \,\hat{\!\bld{\theta}}_{k\!+\!1} \ \qquad \text{and} \ \qquad \,\dot{\!\bld{\theta}}_{k\!+\!1} \ = \ \,\dot{\hat{\!\bld{\theta}}}_{k\!+\!1}  \label{eq:res_b} \\
\bld{x}_{{\scriptsize \mbox{min}}} & \leq & \bld{x}_{k} \ \ \leq \ \ \ \bld{x}_{{\scriptsize \mbox{max}}} \qquad k=0 \cdots N \label{eq:res_c} \\
\dot{\bld{x}}_{{\scriptsize \mbox{min}}} & \leq & \dot{\bld{x}}_{k} \ \ \leq \ \ \ \dot{\bld{x}}_{{\scriptsize \mbox{max}}} \qquad k=0 \cdots N \label{eq:res_d} \\
\bld{\theta}_{{\scriptsize \mbox{min}}} & \leq & \bld{\theta}_k \ \ \leq \ \ \ \bld{\theta}_{{\scriptsize \mbox{max}}} \qquad k=0 \cdots N \label{eq:res_e} \\
\dot{\!\bld{\theta}}_{{\scriptsize \mbox{min}}} & \leq & \ \dot{\!\bld{\theta}}_k \ \ \leq \ \ \ \dot{\!\bld{\theta}}_{{\scriptsize \mbox{max}}} \qquad k=0 \cdots N \label{eq:res_f} \\
\ddot{\!\bld{\theta}}_{{\scriptsize \mbox{min}}} & \leq & \ \ddot{\!\bld{\theta}}_k \ \ \leq \ \ \ \ddot{\!\bld{\theta}}_{{\scriptsize \mbox{max}}} \qquad k=0 \cdots N \label{eq:res_g} \\
\bld{f}\!_{x\,{\scriptsize \mbox{min}}} & \leq & \bld{f}\!_{xk} \ \leq \ \ \ \bld{f}\!_{x\,{\scriptsize \mbox{max}}} \quad k=0 \cdots N\!\!-\!\!1 \label{eq:res_h}
\end{eqnarray}
\label{eq:costf2_}
\end{subequations}
where $J$ is the objective function that combines running and terminal costs. Constraints (\ref{eq:res_a}) force the initial state position and velocity to be current measurements to achieve closed-loop feedback. Restriction (\ref{eq:res_b}) forces the decision-variable states to be the same as the approximated by (\ref{eq:int}), which is what actually creates the multiple-shooting matching. The constraints (\ref{eq:res_c}-\ref{eq:res_d}) bound the robot's cartesian working space in position $\bld{x}_{k}$ and velocity $\dot{\bld{x}}_{k}$. The joint limits are established in position (\ref{eq:res_e}), velocity (\ref{eq:res_f}), acceleration (\ref{eq:res_g}) and effort (\ref{eq:res_h}). The minimum thresholds for joint acceleration are denoted by $\ddot{\!\bld{\theta}}_{{\scriptsize \mbox{min}}}$ and the maximum by $\ddot{\!\bld{\theta}}_{{\scriptsize \mbox{max}}}$.

\section{RT Nonlinear Model Predictive Control}
\label{sec:nmpc}
\begin{figure}[h!] 
	\centering
	\includegraphics[width=1\columnwidth]{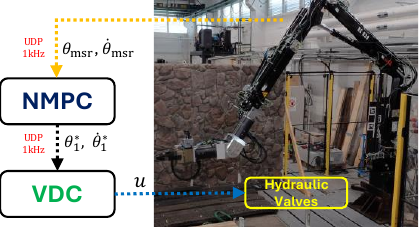}
	\caption{{\footnotesize {\bf Experimental setup.} Angular position $\bld{\theta}_{{\scriptsize \mbox{msr}}}$ and velocity $\,\dot{\!\bld{\theta}}_{{\scriptsize \mbox{msr}}}$ measurements are streamed at 1 kHz. Our NMPC computes an optimal control $\dot{\!\bld{\theta}}^{*}$ every 1 ms to the low-level controller VDC which commands voltage $u$ to the robot's hydraulic valves through the Beckoff system.}}
	\label{fig:overview}
\end{figure}

The experimental setup is depicted in Fig. \ref{fig:overview} where the real heavy-duty manipulator is shown. In order to achieve real-time NMPC, we implement a set of techniques that enable our \textbf{Algorithm \ref{algo:nmpc}} to run in competitive times with full nonlinear dynamics.

For instance, Line 3 of the \textbf{Algorithm \ref{algo:nmpc}} generates an optimized symbolic code for the NLP and its first two partial derivatives. Such code is integrated into a C++ numeric nonlinear solver where the combination of symbolic functions, BFGS and re optimization with warm start generate a NMPC capable to find an optimal solution in less time than the sensors sampling rates. From lines 5 to 8, the solver prepares a suitable warm start solution by allowing a small tolerance in the solver and high number of iterations for getting a refined solution, this takes a random primal solution that respects the constraints. The optimal solution for the decision variable $\bld{z}^*$ is then used as a primal solution of the online routines (lines 10-17). Line 11 read the current measurements via UDP where we implement efficient C++ routines for smart managing of the UDP buffers, this avoids queues and storage of old information and makes our algorithm execution time deterministic which is a required feature for real time implementations. The TCP reference in position and velocity are evaluated at line 12 since we have assumed that the reference trajectories are continuous functions of time, thus they can be evaluated at any instant of time. Line 13 prepares a suitable primal solution by shifting the controls and states of the previous-iteration optimal solution. The solver optimizes at line 14 where only a very small number of iterations of the solver are allowed. This is justified by the slow dynamics of a heavy-duty manipulator, where the NLP doesn't change significantly among consecutive iterations of the NMPC, this allows to take advantage of the sensors high rate, then the decision variable is optimized through iterations of the NMPC rather than iterations of the solver itself \cite{grandia2023perceptive}. If current time is $t_0$ then the optimal motion commands for instant $t_1$ are extracted in line 15 and streamed via UDP to the VDC controller.
	\begin{algorithm}[h!]
		\scriptsize{
			\begin{algorithmic}[1]
            \State \textbf{Offline (one-time setup):}
				\State \mbox{Define} $N$, $\bld{x}^{r}\!(t)$, $\dot{\bld{x}}^{r}\!(t)$, $Q$, $R$, minimum and maximum limits for (\ref{eq:res_c}-\ref{eq:res_h}).
                \State C++ symbolic code generation of the NLP described in \ref{sec:nlp}.
                \State Integrate code into a numeric solver for a real-time execution.
                \State \textbf{Initialization (preparing warm-start solution):}
                \State Receive initial measurements $\bld{\theta}_{{\scriptsize \mbox{msr}}}$ and $\,\dot{\!\bld{\theta}}_{{\scriptsize \mbox{msr}}}$ from sensors via \texttt{UDP}.
                \State Compute reference trajectory $\{\bld{x}^{r}_{k}, \ \dot{\bld{x}}^{r}_{k}\}_{k=1}^{N}$
                \State Call solver with small tolerance and allowing high number of iterations.
                \State \textbf{Online RT NMPC:}
				\While{NMPC is On}
                \State Read current measurements $\bld{\theta}_{{\scriptsize \mbox{msr}}}$ and $\,\dot{\!\bld{\theta}}_{{\scriptsize \mbox{msr}}}$ from sensors via \texttt{UDP}.
                \State Update reference $\{\bld{x}^{r}_{k}, \ \dot{\bld{x}}^{r}_{k}\}_{k=1}^{N}$ where $t_0$ is the current time.
                \State Warm-start solver using shifted previous solution as primal solution.
                \State Solve NLP with fixed-and-small number of iterations.
                \State Extract and stream optimal motion commands $\bld{\theta}^{*}_{1}$ and $\dot{\!\bld{\theta}}^{*}_{1}$ via \texttt{UDP}.
                \State Shift optimal $\bld{z}^{*}$ as primal solution for warm-starting the next iteration.
                \EndWhile
			\end{algorithmic}
		}
		\caption{\small{{\textbf {Real-Time NMPC Execution Pipeline}}}}
		\label{algo:nmpc}
	\end{algorithm}
\section{Experimental Results}
\label{sec:results}
The low-level control system VDC is implemented on a Beckhoff CX2030 industrial PC with a 1 [$ms$] sampling rate as reported in \cite{hejrati2025impact}. The heavy-duty manipulator is actuated through electro hydraulic servo
valves Bosch 4WRPEH10 (100 [$dm^3/min$]) at $\Delta p$ = 3.5 [$MPa$ per notch]. Joint angles are measured using SICK AFS60 absolute encoders (18-bit resolution), and hydraulic pressures are
acquired via Unix 5000 pressure transmitters with a maximum
operating pressure of 25 MPa. The upper-level controller NMPC is running on an industrial-grade computing platform (Nuvo-9160GC) equipped with an Intel Core i9 processor, an NVIDIA RTX 3050 GPU, 32 GB of RAM, and a 1 TB SSD, running a Linux-based operating system with a C++ implementation. The NLP is coded in M{\footnotesize ATLAB} using C{\footnotesize ASADI} for C++ code generation. The NLP and its first order symbolic derivatives are linked to the nonlinear solver I{\footnotesize POPT}. The library B{\footnotesize OOST} is included for managing the UDP connections.

In order to achieve RT execution time for the following experimentation, we set $N=3$ for the circumference and $N=8$ for the spiral-shape trajectory, $\Delta_t = 0.3$ and use Euler step as numeric integrator. I{\footnotesize POPT} solver is enabled to use BFGS for Hessian approximation, its tolerance is set to $1e-6$ and maximum number of iterations as $1$. This last setting allows to perform only one iteration of SQP by NMPC iteration where the NLP solution is optimized thanks to the warm starting and high rate of sensors sampling \cite{grandia2023perceptive}. The weighting matrices are set constant as $R\!=\!\text{diag}(1e-12,1e-12)$, $Q_x \!=\! Q_{x\!N} \!=\! \text{diag}(1e5,1e5)$ and $Q_{\dot{x}} \!=\! Q_{\dot{x}\!N} \!=\! \text{diag}(1e-12,1e-12)$. 

\begin{figure}[h] 
	\centering
	\includegraphics[width=1\columnwidth]{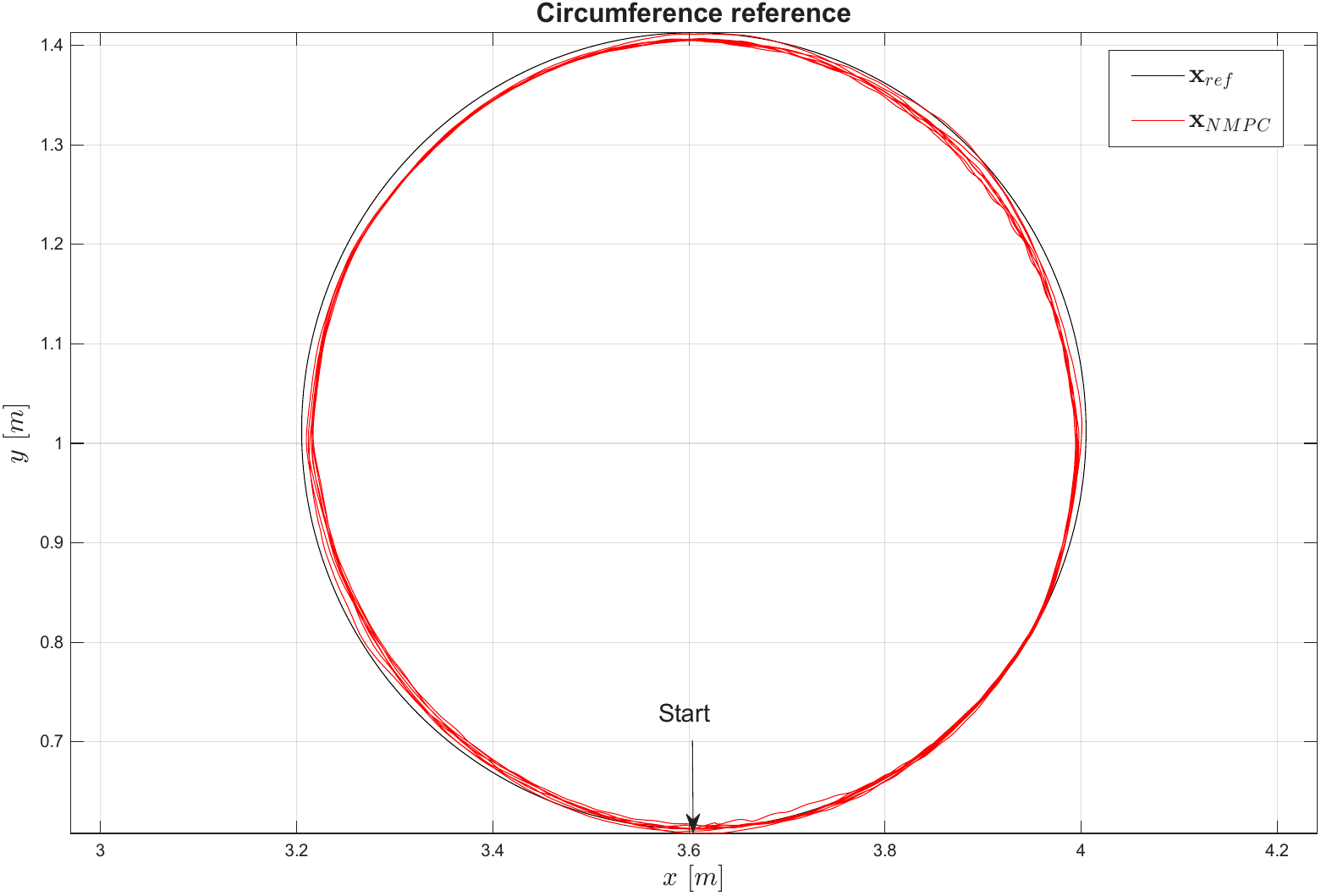}
	\vspace*{-0.3cm}
	\caption{{{\bf Circumference reference trajectory for hydraulic robot's TCP}}}
	\label{fig:Circle_xy}
\end{figure}

\begin{figure}[h] 
	\centering
	\includegraphics[width=1\columnwidth]{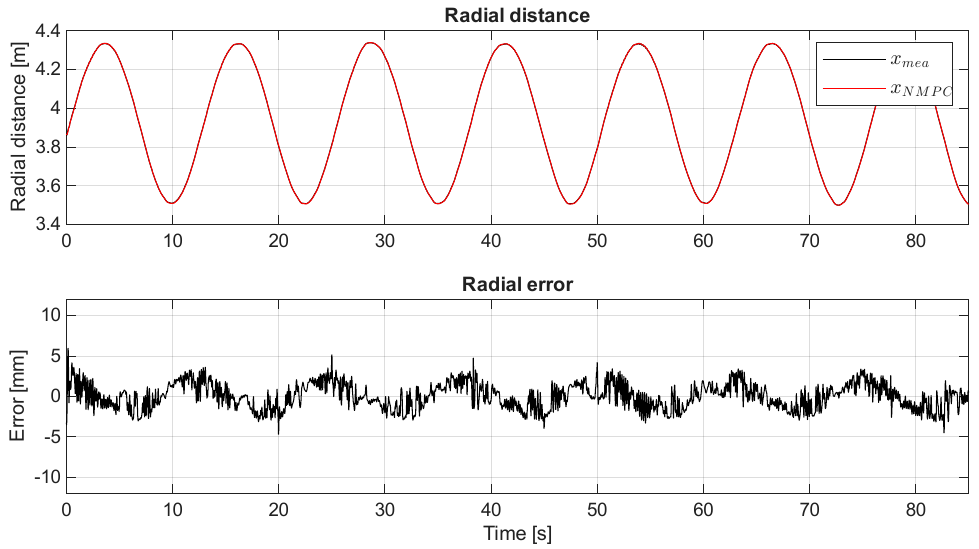}
	\vspace*{-0.6cm}
	\caption{{{\bf Low-level control system tracking performance with circumference reference.}}}
	\label{fig:Circle_vdc}
\end{figure}

\begin{figure}[h] 
	\centering
	\includegraphics[width=1\columnwidth]{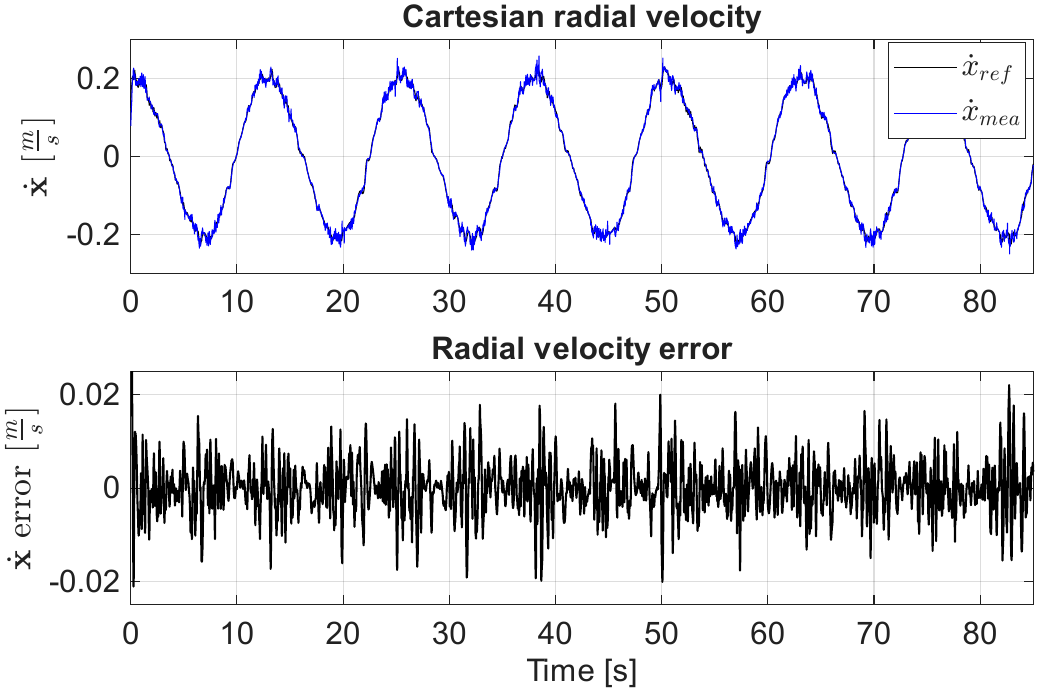}
	\vspace{-0.6cm}
	\caption{{{\bf Cartesian space radial velocity for Circumference trajectory.}}}
	\label{fig:Circle_vels}
\end{figure}

First, the effectiveness of the proposed NMPC framework in accurately tracking trajectories and achieving real-time computation was demonstrated. In this experiment, the hydraulic robot's TCP, shown in Fig. \ref{fig:overview}, was controlled to follow a circumference with a radius of 0.4 m and a tangential velocity of 0.5 m/s in \textit{xy}-plane. The reference trajectory used is shown in Fig.~\ref{fig:Circle_xy}, where $x_{ref}$ denotes the human-defined reference trajectory and $x_{NMPC}$ represents the trajectory optimized by the NMPC framework for the low-level controller. The RMS errors between the reference trajectory and the NMPC-optimized trajectory, as shown in Fig.~\ref{fig:Circle_xy}, along with the corresponding joint space references, were
\begin{itemize}
\item RMS of cartesian position error = 0.021258 \textit{m}
\item RMS of cartesian velocity error = 0.009776 \textit{m/s}
\item RMS of the joint position error = 0.025287 \textit{rad} and 0.033505 \textit{rad} for the Lift and Tilt joint
\item RMS of the joint velocity error = 0.035976 \textit{rad/s} and 0.044383 \textit{rad/s} for the Lift and Tilt joint
\end{itemize}
%
The performance of the low-level control system with real-time references from NMPC is presented in Fig.~ \ref{fig:Circle_vdc} and Fig.~\ref{fig:Circle_vels}. As Fig.~\ref{fig:Circle_vdc} shows, the average radial tracking error during the control sequence is less than 5 mm. Also, the radial velocity error in low-level controller were less than 0.02 m/s, as presented in Fig.~\ref{fig:Circle_vels}. Such a performance is due to the excellent behavior of the NMPC to make the desired, human-defined trajectory compliant with the system capabilities and limitations. 

%
%
%
%

\begin{figure} 
	\centering
	\includegraphics[width=0.9\columnwidth]{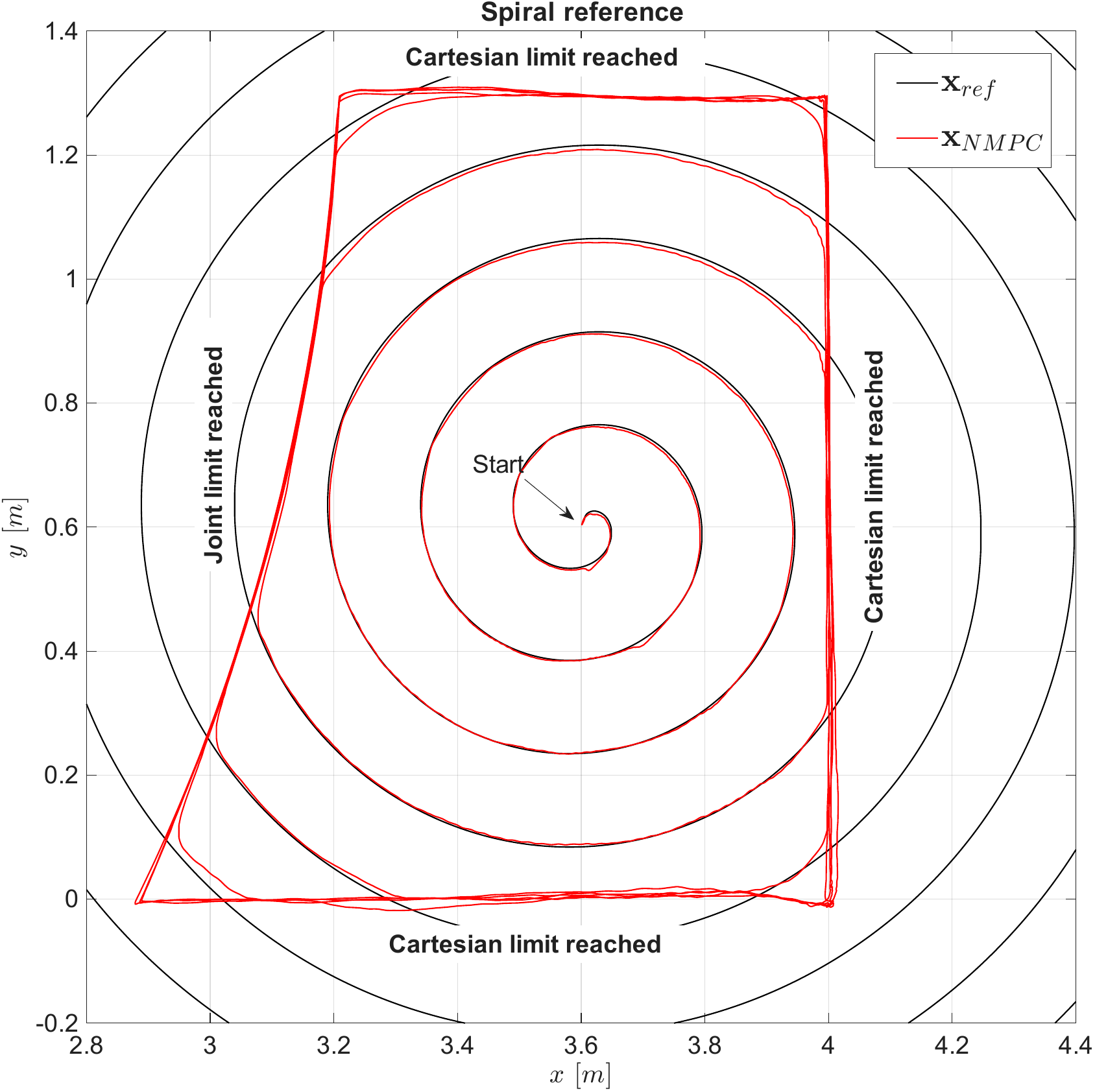}
    \vspace*{-0.3cm}
	\caption{{\footnotesize {\bf Spiral reference trajectory with increasing tangential speed of TCP.}}}
	\label{fig:88_doble}
\end{figure}

\begin{figure*}[t] 
	\centering
	\includegraphics[width=\textwidth]{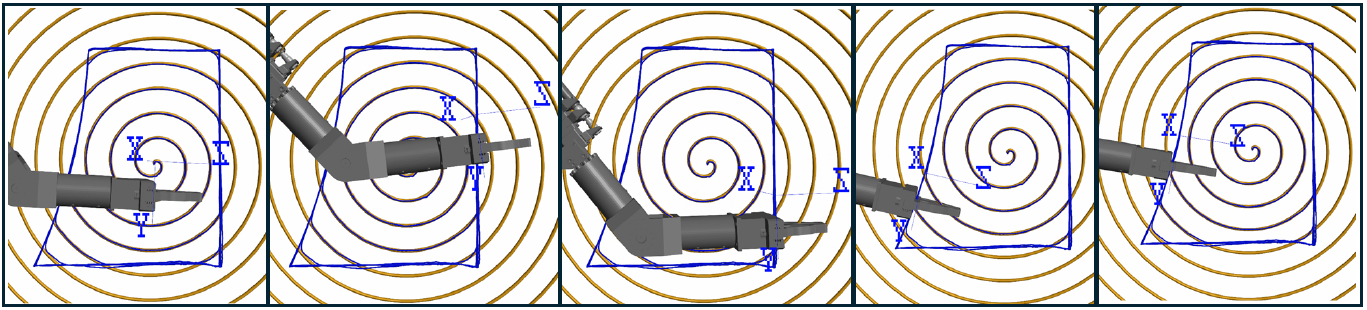}
    \vspace*{-0.3cm}
	\caption{{\bf NMPC considers both Cartesian and joint space constraints during real-time trajectory generation}}
	\label{fig:Limits}
\end{figure*}

To evaluate the effectiveness of the proposed NMPC framework in handling Cartesian and joint space constraints during trajectory generation, a spiral-shaped trajectory with increasing tangential speed of the TCP is illustrated in Fig. \ref{fig:88_doble}. The limits of the working space were set as $\bld{x}_{{\scriptsize \mbox{min}}} \!=\! [0, 0]$, $\bld{x}_{{\scriptsize \mbox{max}}} \!=\! [4.0, 1.3]$, $\bld{\theta}_{{\scriptsize \mbox{min}}} \!=\! [-0.2,-1.9]$, $\bld{\theta}_{{\scriptsize \mbox{max}}} \!=\! [1,-0.7]$, $\dot{\!\bld{\theta}}_{{\scriptsize \mbox{min}}} \!=\! [-0.35, -0.35]$, $\dot{\!\bld{\theta}}_{{\scriptsize \mbox{max}}} \!=\! [0.35,0.35]$, $\ddot{\!\bld{\theta}}_{{\scriptsize \mbox{min}}} \!=\! [-0.3, -0.3]$, $\ddot{\!\bld{\theta}}_{{\scriptsize \mbox{max}}} \!=\! [0.3, 0.3]$, $\bld{f}\!_{x\,{\scriptsize \mbox{min}}} \!=\! [-5e4, -5e4]$, and $\bld{f}\!_{x\,{\scriptsize \mbox{max}}} \!=\! [5e4, 5e4]$. 
As illustrated in Fig.~\ref{fig:Limits}, the spiral reference starts from the initial position. The TCP follows the spiral path until it reaches the first Cartesian space limit along the x-axis. At that point, the TCP moves along the boundary of the limit. Then it continues to follow the spiral reference until a joint limit is reached. Afterward, the joint limits begin to influence the Cartesian reference. Finally, Cartesian and configurational limits were reached in all directions in the plane, as Fig. \ref{fig:88_doble} shows. 

\begin{figure} 
	\centering
	\includegraphics[width=1\columnwidth]{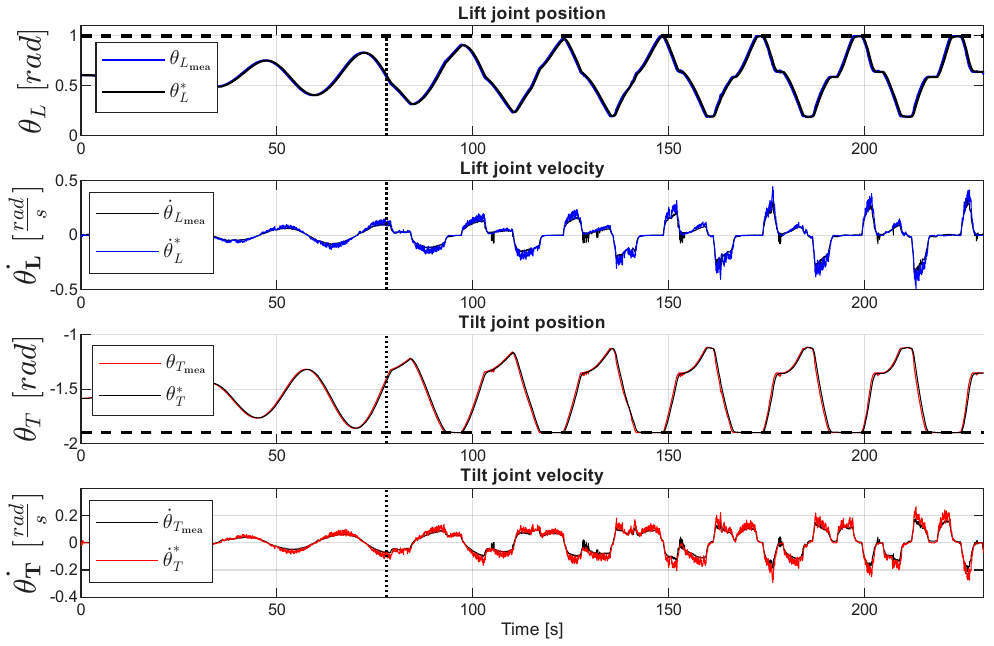}
    \vspace*{-0.3cm}
	\caption{{ {\bf NMPC reference joint positions and velocities.} The joint tracking in position and velocity correspond to the VDC low-level controller. The horizontal dotted line represents the instant of time, $t=78 s$, where the limits are exceeded. The vertical dashed line represents the joint limit.}}
	\label{fig:88_doble_out}
\end{figure}

\begin{figure} 
	\centering
	\includegraphics[width=1\columnwidth]{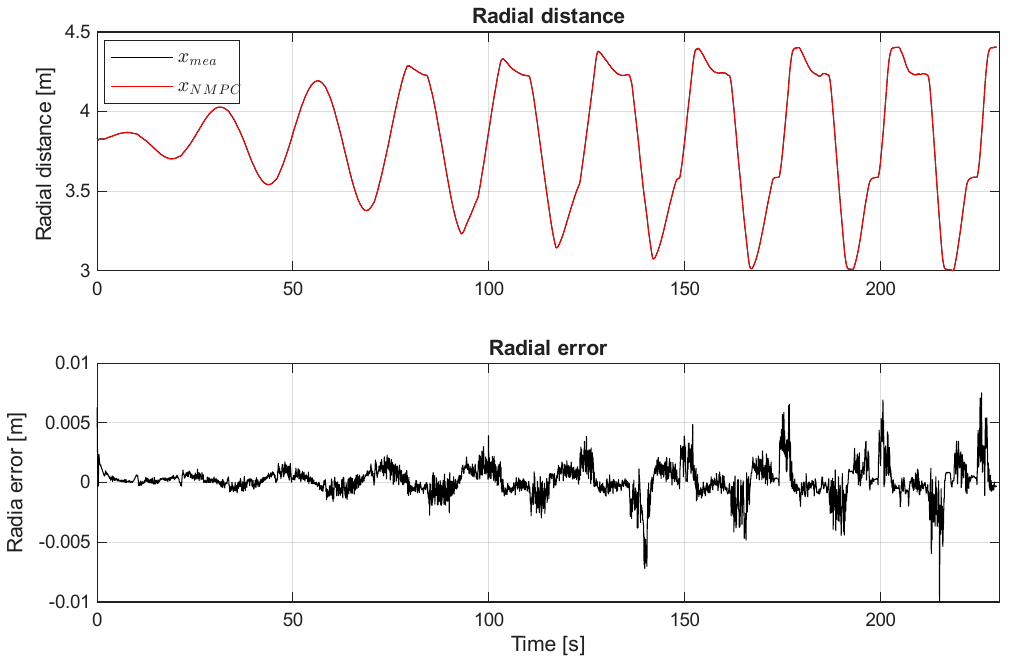}
    \vspace*{-0.3cm}
	\caption{{ {\bf Radial distance of hydraulic robot's TCP}}}
	\label{fig:88_VDC_out}
\end{figure}

The corresponding real-time references for Lift and Tilt joint positions and velocities are presented in Fig.~\ref{fig:88_doble_out}. For spiral reference, the RMS errors between the reference trajectory and the NMPC-optimized trajectory
the RMS errors for Cartesian reference, in Fig.~\ref{fig:88_doble} and joint references in Fig.~\ref{fig:88_doble_out} were  
\begin{itemize}
\item RMS of the Cartesian position error = 0.006944 \textit{m}
\item RMS of the Cartesian velocity error = 0.006474 \textit{m/s}
\item RMS of joint position error = 0.019402 \textit{rad} and 0.0250045 \textit{rad}
\item RMS of joint velocity error = 0.021604 \textit{rad/s} and 0.0246659 \textit{rad/s}
\end{itemize}
The control performance of the VDC with NMPC real-time references is presented in Fig.~\ref{fig:88_VDC_out}. As Fig.~\ref{fig:88_VDC_out} shows, the average radial tracking error during the control sequence was less than 5 mm.


\section{Conclusions}
\label{sec:conclusion}
We have presented a novel NMPC framework that operates on the complete nonlinear dynamics of HHMs to achieve high-precision trajectory tracking. The proposed controller ensures deterministic and bounded computation time, enabled by a combination of solver warm-starting, high-frequency sensor updates (1~kHz), and a limited number of optimization iterations—allowing the optimal solution to be computed within the interval of the fastest sensor measurements. We have evaluated the effectiveness of the NMPC framework through experimental validation on two representative trajectories. First, as shown in Fig.~\ref{fig:Circle_vdc}, the controller achieves high-accuracy end-effector tracking during a circular motion task. Second, Fig.~\ref{fig:88_doble} highlights the ability of the proposed NMPC to enforce actuator and system constraints in both joint and Cartesian spaces during aggressive, fast-paced maneuvers. These results confirm that our framework not only meets real-time execution requirements, but also enables safety-aware control for complex hydraulic systems operating under strict physical constraints.

\bibliography{root,MD_Ref}

\begin{thebibliography}{10}

\bibitem{hejrati2025orchestrated}
Mahdi Hejrati and Jouni Mattila.
\newblock Orchestrated robust controller for precision control of heavy-duty hydraulic manipulators.
\newblock {\em IEEE Transactions on Automation Science and Engineering}, 2025.

\bibitem{hejrati2025robust}
Mahdi Hejrati, Pauli Mustalahti, and Jouni Mattila.
\newblock Robust immersive bilateral teleoperation of beyond-human-scale systems with enhanced transparency and sense of embodiment.
\newblock {\em arXiv preprint arXiv:2505.14486}, 2025.

\bibitem{mattila2017survey}
Jouni Mattila, Janne Koivum{\"a}ki, Darwin~G Caldwell, and Claudio Semini.
\newblock A survey on control of hydraulic robotic manipulators with projection to future trends.
\newblock {\em iEeE/ASME Transactions on Mechatronics}, 22(2):669--680, 2017.

\bibitem{ortiz2014increasing}
Daniel Ortiz~Morales, Simon Westerberg, Pedro~X La~Hera, Uwe Mettin, Leonid Freidovich, and Anton~S Shiriaev.
\newblock Increasing the level of automation in the forestry logging process with crane trajectory planning and control.
\newblock {\em Journal of Field Robotics}, 31(3):343--363, 2014.

\bibitem{feng2018robotic}
Hao Feng, Chen-Bo Yin, Wen-wen Weng, Wei Ma, Jun-jing Zhou, Wen-hua Jia, and Zi-li Zhang.
\newblock Robotic excavator trajectory control using an improved ga based pid controller.
\newblock {\em Mechanical Systems and Signal Processing}, 105:153--168, 2018.

\bibitem{egli2022general}
Pascal Egli and Marco Hutter.
\newblock A general approach for the automation of hydraulic excavator arms using reinforcement learning.
\newblock {\em IEEE robotics and automation letters}, 7(2):5679--5686, 2022.

\bibitem{taheri2024end}
Abdolreza Taheri, Amy Rankka, Pelle Gustafsson, Joni Pajarinen, and Reza Ghabcheloo.
\newblock End-effector cartesian velocity control for redundant loader cranes using reinforcement learning.
\newblock {\em IEEE Transactions on Robotics}, 2024.

\bibitem{zhu2010virtual}
Wen-Hong Zhu.
\newblock {\em Virtual decomposition control: toward hyper degrees of freedom robots}, volume~60.
\newblock Springer Science \& Business Media, 2010.

\bibitem{koivumaki2019energy}
Janne Koivum{\"a}ki, Wen-Hong Zhu, and Jouni Mattila.
\newblock Energy-efficient and high-precision control of hydraulic robots.
\newblock {\em Control Engineering Practice}, 85:176--193, 2019.

\bibitem{helian2023constrained}
Bobo Helian, Zheng Chen, and Bin Yao.
\newblock Constrained motion control of an electro-hydraulic actuator under multiple time-varying constraints.
\newblock {\em IEEE Transactions on Industrial Informatics}, 19(12):11878--11888, 2023.

\bibitem{xu2022eso}
Zhangbao Xu, Guoliang Qi, Qingyun Liu, and Jianyong Yao.
\newblock Eso-based adaptive full state constraint control of uncertain systems and its application to hydraulic servo systems.
\newblock {\em Mechanical Systems and Signal Processing}, 167:108560, 2022.

\bibitem{xu2022extended}
Zhangbao Xu, Wenxiang Deng, Hao Shen, and Jianyong Yao.
\newblock Extended-state-observer-based adaptive prescribed performance control for hydraulic systems with full-state constraints.
\newblock {\em IEEE/ASME Transactions on Mechatronics}, 27(6):5615--5625, 2022.

\bibitem{garcia1989model}
Carlos~E Garcia, David~M Prett, and Manfred Morari.
\newblock Model predictive control: Theory and practice—a survey.
\newblock {\em Automatica}, 25(3):335--348, 1989.

\bibitem{diehl2009efficient}
Moritz Diehl, Hans~Joachim Ferreau, and Niels Haverbeke.
\newblock Efficient numerical methods for nonlinear mpc and moving horizon estimation.
\newblock {\em Nonlinear model predictive control: towards new challenging applications}, pages 391--417, 2009.

\bibitem{varga2019model}
Balint Varga, Selina Meier, Stefan Schwab, and S{\"o}ren Hohmann.
\newblock Model predictive control and trajectory optimization of large vehicle-manipulators.
\newblock In {\em 2019 IEEE International Conference on Mechatronics (ICM)}, volume~1, pages 60--66. IEEE, 2019.

\bibitem{ma2024data}
Dexian Ma and Bo~Zhou.
\newblock Data-driven multi-step nonlinear model predictive control for industrial heavy load hydraulic robot.
\newblock {\em arXiv preprint arXiv:2411.13859}, 2024.

\bibitem{kalmari2014nonlinear}
Jouko Kalmari, Juha Backman, and Arto Visala.
\newblock Nonlinear model predictive control of hydraulic forestry crane with automatic sway damping.
\newblock {\em Computers and Electronics in Agriculture}, 109:36--45, 2014.

\bibitem{kleff2021high}
S{\'e}bastien Kleff, Avadesh Meduri, Rohan Budhiraja, Nicolas Mansard, and Ludovic Righetti.
\newblock High-frequency nonlinear model predictive control of a manipulator.
\newblock In {\em 2021 IEEE International Conference on Robotics and Automation (ICRA)}, pages 7330--7336. IEEE, 2021.

\bibitem{bock1984multiple}
Hans~Georg Bock and Karl-Josef Plitt.
\newblock A multiple shooting algorithm for direct solution of optimal control problems.
\newblock {\em IFAC Proceedings Volumes}, 17(2):1603--1608, 1984.

\bibitem{diehl2006fast}
Moritz Diehl, Hans~Georg Bock, Holger Diedam, and P-B Wieber.
\newblock Fast direct multiple shooting algorithms for optimal robot control.
\newblock {\em Fast motions in biomechanics and robotics: optimization and feedback control}, pages 65--93, 2006.

\bibitem{alvaro2024analytical}
Paz Alvaro and Jouni Mattila.
\newblock Analytical forward dynamics modeling of linearly actuated heavy-duty parallel-serial manipulators.
\newblock {\em arXiv preprint arXiv:2403.08524}, 2024.

\bibitem{petrovic2022mathematical}
Goran~R Petrovi{\'c} and Jouni Mattila.
\newblock Mathematical modelling and virtual decomposition control of heavy-duty parallel--serial hydraulic manipulators.
\newblock {\em Mechanism and Machine Theory}, 170:104680, 2022.

\bibitem{Bib:Betts}
J.-T. Betts.
\newblock {\em Practical Methods for Optimal Control and Estimation Using Nonlinear Programming}.
\newblock Advances in Design and Control. SIAM, 2nd edition, 2010.

\bibitem{grandia2023perceptive}
Ruben Grandia, Fabian Jenelten, Shaohui Yang, Farbod Farshidian, and Marco Hutter.
\newblock Perceptive locomotion through nonlinear model-predictive control.
\newblock {\em IEEE Transactions on Robotics}, 39(5):3402--3421, 2023.

\bibitem{hejrati2025impact}
Mahdi Hejrati and Jouni Mattila.
\newblock Impact-resilient orchestrated robust controller for heavy-duty hydraulic manipulators.
\newblock {\em IEEE/ASME Transactions on Mechatronics}, 2025.

\end{thebibliography}
\bibliographystyle{unsrt}

\end{document}